\documentclass[11pt]{article}
\usepackage[margin=1in]{geometry}
\usepackage{amsmath,amssymb,amsthm}
\usepackage{booktabs}
\usepackage{graphicx}
\usepackage{multirow}
\usepackage{microtype}
\usepackage[numbers,sort&compress]{natbib}
\usepackage[colorlinks=true,linkcolor=blue,citecolor=blue,urlcolor=blue]{hyperref}
\usepackage{xcolor}

\newcommand{\method}{HyperWorld}

\title{\method: Hypergraph-Structured State Serialization\\
Improves Learned Textual World Models}

\author{%
Yun-Jian Zhang\textsuperscript{1}\quad
Chen-Wei Liang\textsuperscript{2}\quad
Tian-Yi Zhang\textsuperscript{3}\quad
Jian Ding\textsuperscript{1}\quad
Yi-Lun Wu\textsuperscript{1}\\[3pt]
Ao-Bo Li\textsuperscript{1}\quad
Wei-Cong Su\textsuperscript{1}\quad
Saifullah\textsuperscript{1}\quad
Hong-Yu An\textsuperscript{4}\quad
Mu-Jiang-Shan Wang\textsuperscript{1,5,*}\\[6pt]
\textsuperscript{1}\,Shenzhen Kaihong Digital Industry Development Co., Ltd.\\
\texttt{\{zhangyunjian,wuyilun,Liaobo\}@kaihong.com}\\
\texttt{\{suweicong,saifullah\}@kaihong.com}; \texttt{dj19@tsinghua.org.cn}\\[3pt]
\textsuperscript{2}\,School of Mathematics and Statistics, Faculty of Science, University of New South Wales,\\
Sydney, NSW 2052, Australia; \texttt{z5537371@ad.unsw.edu.au}\\[3pt]
\textsuperscript{3}\,College of Computer Science and Technology, Zhejiang University,\\
38 Zheda Road, Hangzhou 310027, China; \texttt{tianyizhang0213@zju.edu.cn}\\[3pt]
\textsuperscript{4}\,State Key Laboratory of Internet of Things for Smart City (SKL-IoTSC),\\
University of Macau, Macau; \texttt{yc48116@connect.um.edu.mo}\\[3pt]
\textsuperscript{5}\,Shenzhen Institute of Advanced Technology, Chinese Academy of Sciences,\\
Shenzhen, China; \texttt{mjs.wang@siat.ac.cn}\\[3pt]
\textsuperscript{*}\,Corresponding author: \texttt{mjs.wang@siat.ac.cn}
}

\date{}

\begin{document}
\maketitle

\begin{abstract}
World models, which predict how an environment evolves under actions, are
increasingly used to equip language-model agents with the ability to plan
before acting. In text environments, a world model must learn symbolic
dynamics from serialized descriptions of the state, yet how the
\emph{structure} of this serialization affects learning remains largely
unexamined: prior work serializes states either as flat text or as pairwise
relational triples, both of which fragment the joint, higher-order structure
of environment states. We present \method, a systematic study of state
serialization structure for learning textual world models with small language
models. We compare four information-equivalent representations of the same
ground-truth symbolic state---raw observations, independent sentences,
pairwise triples, and entity-centred \emph{hyperedge units} that jointly
describe multiple entities and their relations---under an identical
effect-prediction objective: given a state and an action, predict the
symbolic effects or judge the action infeasible. Across model scales (0.5B--3B) and data budgets, hyperedge serialization
improves effect prediction most clearly at 0.5B--1.5B and on
out-of-distribution (OOD) test worlds, where it retains more of its
in-distribution accuracy than raw text and, in most settings, outperforms
pairwise triples. At 3B, larger models narrow the gap---triples even match or
slightly exceed hyperedge EM on IID data---but hyperedge grouping still
achieves the strongest OOD fact F1. Raw observations support strong
feasibility detection but weak effect prediction; hyperedge units recover
near-raw feasibility while improving effects, yielding the best overall
trade-off at small-to-medium scale. A greedy planner using the hyperedge
world model also attains the highest success rate among the representations
we test. These results suggest that higher-order state structure is a cheap
inductive bias for learned symbolic world models, especially when capacity is
limited or test worlds differ from training.
\end{abstract}

\section{Introduction}
\label{sec:intro}

World models---internal predictive models of environment dynamics---are a
central ingredient of deliberative agents: they allow an agent to
\emph{imagine} the consequences of candidate actions before executing them,
enabling lookahead planning, risk assessment, and sample-efficient policy
learning~\citep{ha2018world,hafner2023dreamerv3,lecun2022path}. Recent
large-scale systems extend this paradigm to pixels and video
\citep{bruce2024genie,agarwal2025cosmos}, while a parallel line of work
equips language-model agents with \emph{textual} world models that predict
environment feedback or symbolic state changes in interactive text
environments~\citep{ammanabrolu2021worldformer,chae2025web,gu2024proposes}.

A textual world model must consume a serialized description of the current
state. Existing approaches serialize states either as raw textual
observations or as a set of pairwise relational triples
\citep{ammanabrolu2021worldformer,anokhin2024arigraph,xu2025gwm}---the same
pairwise decomposition used by knowledge graphs. However, environment states
are intrinsically \emph{higher-order}: the conditions and effects of an
action typically involve several entities jointly. Unlocking a chest, for
instance, simultaneously concerns the player's location, the key in the
inventory, the chest's lock state, and the key--lock correspondence.
Pairwise decomposition scatters this joint structure across disconnected
triples, and flat text scatters it across sentences, leaving the model to
re-assemble the relevant context from fragments at every prediction step.This concern is conceptually related to classical graph-theoretic
studies of reachability and navigation, where orientation,
Hamiltonicity, and path embedding provide structural ways to reason about
feasible transitions and robust routes in discrete state spaces
\citep{zhao2017algorithm,mu2010ordered,wang2011embedding}.
Here we ask an analogous question for textual world models: whether the
serialization of a symbolic state exposes the joint transition structure
needed for learning action effects.
Hypergraphs, whose hyperedges connect arbitrary numbers of vertices, are a
natural formalism for such joint structure, and have recently proven useful
for retrieval-augmented generation~\citep{luo2025hypergraphrag} and knowledge
representation. Whether higher-order structure helps a model \emph{learn
dynamics} is, to our knowledge, unexplored.

This paper asks a deliberately focused question:
\begin{quote}
\emph{Holding information content fixed, does hypergraph-structured state
serialization improve a small language model's ability to learn textual world
dynamics?}
\end{quote}

To answer it we build \method, a controlled experimental framework on
procedurally generated TextWorld games~\citep{cote2018textworld}. From
ground-truth symbolic states we derive four serializations with identical
information content but different structure: (i) raw observations, (ii)
independent natural-language sentences, (iii) pairwise triples, and (iv)
entity-centred hyperedge units that jointly render an entity's location,
internal state, contents, and key bindings as a single unit. A small language
model (Qwen2.5, 0.5B--3B)~\citep{qwen2.5} is LoRA-fine-tuned
\citep{hu2022lora} as a world model that maps a serialized state and an
action to symbolic effects (\texttt{ADD}/\texttt{REMOVE} facts) or an
\texttt{INFEASIBLE} verdict. Because targets are identical across
serializations, any performance difference is attributable to input structure
alone.

Our contributions are threefold:
\begin{itemize}
\item \textbf{A controlled testbed} for studying state-serialization
structure in learned textual world models, with information-equivalent
flat, pairwise, and hypergraph renderings of ground-truth symbolic states,
spanning IID and out-of-distribution splits.
\item \textbf{Empirical evidence} that hyperedge serialization improves
effect prediction and OOD robustness at 0.5B--1.5B, recovers much of raw
text's feasibility signal that flat fact renderings lose at 0.5B, and
remains competitive at 3B where triples are strong on IID metrics.
\item \textbf{A planning demonstration} showing that, under the same greedy
search procedure, the hyperedge world model attains a higher live-game
success rate than sentence- or triple-based world models.
\end{itemize}

\section{Related Work}
\label{sec:related}

\paragraph{World models.}
Learning predictive models of environment dynamics has a long history in
model-based reinforcement learning~\citep{ha2018world,hafner2023dreamerv3}.
Recent foundation-scale world models generate interactive visual
environments~\citep{bruce2024genie,agarwal2025cosmos}, while
JEPA-style architectures learn latent dynamics without pixel
reconstruction~\citep{lecun2022path,assran2025vjepa2}. Our work concerns a
complementary, low-compute regime: small language models learning
\emph{symbolic} dynamics of text environments.

\paragraph{Textual world models for LLM agents.}
Equipping LLM agents with world models improves decision making by allowing
lookahead before execution~\citep{chae2025web,gu2024proposes,liu2026itp}.
Worldformer~\citep{ammanabrolu2021worldformer} predicts knowledge-graph
transitions in text games; AriGraph~\citep{anokhin2024arigraph} maintains a
memory graph for agent reasoning; Graph World Models~\citep{xu2025gwm}
formalize world models over graph-structured states. StateFactory
\citep{statefactory2026} factorizes observations into object--attribute
hierarchies for reward prediction. All of these rely on flat text or pairwise
relations; none examines higher-order serialization structure, which is our
focus.

\paragraph{Graph-theoretic robustness and structural recoverability.}
A related discrete perspective comes from graph-theoretic studies of
network robustness and diagnosability. Diagnosability asks whether faulty
components can be identified from comparison information, while edge
connectivity measures the tolerance of a networked structure to link
failures \citep{wang2016diagnosability,wang2018edge}. Although these
notions differ from learned state serialization, they reflect a common
theme: useful representations should preserve enough structure to support
recovery, robustness, and reliable reasoning under perturbations.

\paragraph{Hypergraphs for language systems.}
Hypergraph representations capture $n$-ary associations that pairwise graphs
fragment, with demonstrated benefits in retrieval-augmented generation
\citep{luo2025hypergraphrag} and multi-hop question answering. We transfer
this insight from knowledge retrieval to \emph{dynamics learning}, and test
whether hyperedge-grouped states make forward simulation easier to learn.

\begin{figure}[t]
\centering
\includegraphics[width=\linewidth]{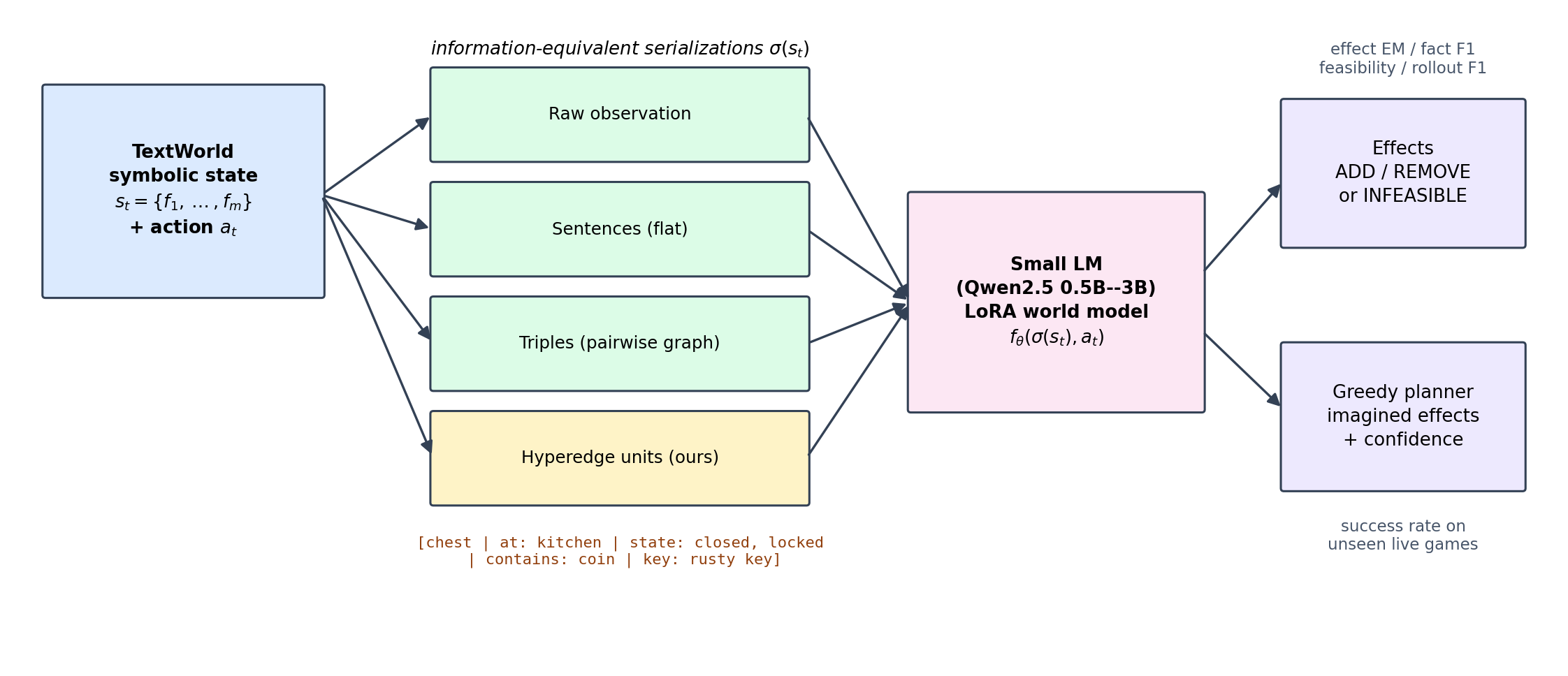}
\caption{Overview of \method. The same ground-truth symbolic state is
rendered by four information-equivalent serializers; a LoRA-fine-tuned small
LM learns to predict symbolic action effects (or infeasibility) from each
rendering. Learned dynamics are evaluated by single-step and rollout metrics
and by a downstream world-model-guided planner on live games.}
\label{fig:overview}
\end{figure}

\section{Method}
\label{sec:method}

\subsection{Problem Setup}
\label{sec:setup}

We model a text environment as a partially observable MDP whose underlying
state is a set of ground facts $s_t = \{f_1, \dots, f_m\}$, where each fact
$f = p(e_1, \dots, e_k)$ applies a predicate $p$ to entities $e_i$ (e.g.,
$\mathrm{in}(\text{key}, \text{chest})$, $\mathrm{locked}(\text{chest})$).
Executing action $a_t$ yields the successor state
$s_{t+1} = (s_t \setminus R_t) \cup A_t$, where $A_t$ and $R_t$ are the added
and removed fact sets. Actions outside the admissible set leave the state
unchanged.

A \emph{textual world model} is a function
\begin{equation}
f_\theta\bigl(\sigma(s_t),\, a_t\bigr) \;\longrightarrow\;
\begin{cases}
(A_t,\, R_t) & \text{if } a_t \text{ is admissible in } s_t,\\
\texttt{INFEASIBLE} & \text{otherwise,}
\end{cases}
\label{eq:wm}
\end{equation}
where $\sigma(\cdot)$ is a \emph{serializer} that renders the symbolic state
as text. The output vocabulary (canonical fact strings) is fixed across
serializers, so the choice of $\sigma$ is the only manipulated variable.

\subsection{Information-Equivalent State Serializations}
\label{sec:serializers}

Given the same fact set $s$, we compare four serializers
(Figure~\ref{fig:reprs}):

\begin{figure}[t]
\centering
\small
\begin{tabular}{p{0.46\textwidth}p{0.46\textwidth}}
\toprule
\textbf{Sentences} ($\sigma_{\mathrm{sent}}$) & \textbf{Triples} ($\sigma_{\mathrm{tri}}$) \\
\midrule
\begin{minipage}[t]{\linewidth}\ttfamily\scriptsize
player is at kitchen.\\
rusty key is in inventory.\\
chest is at kitchen.\\
chest is closed.\\
chest is locked.\\
gold coin is in chest.\\
rusty key matches chest.\\
pantry is west of kitchen.
\end{minipage}
&
\begin{minipage}[t]{\linewidth}\ttfamily\scriptsize
(player, at, kitchen)\\
(rusty key, in, inventory)\\
(chest, at, kitchen)\\
(chest, is, closed)\\
(chest, is, locked)\\
(gold coin, in, chest)\\
(rusty key, match, chest)\\
(pantry, west\_of, kitchen)
\end{minipage}
\\[2pt]
\midrule
\textbf{Hyper} ($\sigma_{\mathrm{hyp}}$, ours) & \textbf{Target (identical for all)} \\
\midrule
\begin{minipage}[t]{\linewidth}\ttfamily\scriptsize
[player | at: kitchen | holds: rusty key]\\
{[}chest | at: kitchen | state: closed, locked | contains: gold coin | key: rusty key]\\
{[}kitchen | contains: chest | exits: west -> pantry]
\end{minipage}
&
\begin{minipage}[t]{\linewidth}\ttfamily\scriptsize
ACTION: unlock chest with rusty key\\
EFFECTS: ADD: none |\\
\hspace*{1em}REMOVE: locked(chest)
\end{minipage}
\\
\bottomrule
\end{tabular}
\caption{Information-equivalent serializations of the same symbolic state.
The hyperedge form groups all facts relevant to an entity into a single
$n$-ary unit, so the joint precondition of an action (key held, chest locked,
key--lock match, co-location) appears in one place rather than scattered
across lines.}
\label{fig:reprs}
\end{figure}

\textbf{Raw} ($\sigma_{\mathrm{raw}}$): the textual observation produced by
the environment (room description and inventory). This is what a purely
text-based world model sees.

\textbf{Sentences} ($\sigma_{\mathrm{sent}}$): each fact rendered as an
independent natural-language sentence (``The key is in the chest.''). A flat
control condition that carries exactly the facts, with no relational
structure.

\textbf{Triples} ($\sigma_{\mathrm{tri}}$): each fact rendered as a binary
triple $(\text{head}, \text{relation}, \text{tail})$; unary predicates become
$(\text{entity}, \texttt{is}, \text{predicate})$. This is the pairwise-graph
representation used by graph-based world models
\citep{ammanabrolu2021worldformer,anokhin2024arigraph,xu2025gwm}.

\textbf{Hyper} ($\sigma_{\mathrm{hyp}}$, ours): facts are grouped into
entity-centred $n$-ary units by a deterministic procedure. Each unit is a
hyperedge over several entities, rendered on one line:
\begin{center}
\texttt{[chest | at: kitchen | state: closed, locked | contains: coin | key: rusty key]}
\end{center}
The grouping covers the player (location and inventory), every object
(location, unary states, contents, key bindings), and room connectivity
(exits); a catch-all unit preserves any remaining facts, keeping the
rendering lossless.

All fact-based serializers ($\sigma_{\mathrm{sent}}, \sigma_{\mathrm{tri}},
\sigma_{\mathrm{hyp}}$) are bijective renderings of the same fact set:
they contain identical information and differ only in how facts are grouped
on the page. This isolates \emph{structure} as the experimental variable.

\subsection{World-Model Learning}
\label{sec:training}

Transitions are collected from procedurally generated games by mixing
goal-directed walkthrough trajectories with $\epsilon$-random branches, which
yields diverse on-path and off-path dynamics. Infeasible actions are sampled
from syntactically valid commands that are not admissible in the current
state, providing negatives for feasibility learning.

The world model is a pretrained decoder-only LM fine-tuned with LoRA on the
unified objective of Eq.~\eqref{eq:wm}, serialized as
\begin{center}
\texttt{STATE: $\sigma(s_t)$\ \ ACTION: $a_t$\ \ EFFECTS: ADD: \dots\ |\ REMOVE: \dots}
\end{center}
with cross-entropy on the target tokens only. Identical targets,
hyperparameters, and data across serializers guarantee a controlled
comparison.

\subsection{World-Model-Guided Planning}
\label{sec:planning}

To test whether better dynamics translate into better behaviour, we use the
learned world model inside a greedy planner on live, unseen games. At each
step, for every admissible command $a$, the planner queries the world model
for imagined effects $(\hat A, \hat R)$ and scores
\begin{equation}
\mathrm{score}(a) = w_g\,\Delta_{\mathrm{goal}}(a) + w_f\,\mathbb{1}[\hat{\text{feasible}}]
 + w_n\,\mathbb{1}[\text{novel imagined state}] + w_c\, c(a),
\label{eq:score}
\end{equation}
where $\Delta_{\mathrm{goal}}$ is the imagined gain in satisfied goal facts,
and $c(a)$ is the model's mean token log-probability, a confidence signal
that down-weights unreliable imaginations. The action with the highest score
is executed. The planner is deliberately simple: it isolates the quality of
the learned dynamics rather than the sophistication of search.

\section{Experiments}
\label{sec:experiments}

\subsection{Setup}
\label{sec:exp-setup}

\paragraph{Environments and data.}
We generate 410 TextWorld games in four splits: train (300 games; 4 rooms, 8
objects, quest length 3), validation (30), test-IID (40 unseen games with
training parameters), and test-OOD (40 games with 8 rooms, 16 objects, quest
length 6). Rolling out one walkthrough trajectory and two
$\epsilon$-random-branching trajectories per game ($\epsilon{=}0.35$, max 25
steps) yields 8{,}375 train / 797 validation / 1{,}058 test-IID / 2{,}001
test-OOD transitions, roughly one third of which are infeasible-action
negatives.

\paragraph{Models and training.}
Qwen2.5-Instruct at 0.5B, 1.5B, and 3B with LoRA ($r{=}16$), 2 epochs,
learning rate $2\times10^{-4}$, identical across conditions. All experiments
run on a single RTX 5090.

\paragraph{Metrics.}
(i) \emph{Feasibility accuracy/F1}: detecting inadmissible actions;
(ii) \emph{Effect exact match}: predicted delta exactly equals the gold
delta; (iii) \emph{Fact F1}: precision/recall over predicted added/removed
facts; (iv) \emph{Rollout state F1}: F1 between the simulated and true fact
sets after $k$ imagined steps; (v) \emph{Planning success rate} on live
games.

\subsection{Main Results}
\label{sec:results-main}

Table~\ref{tab:main} reports effect exact match (EM) and fact-level F1 on
test-IID and test-OOD splits across three model scales (bold = best per
column). \texttt{hyper} is strongest overall at 0.5B and on most OOD
columns, but it is \emph{not} uniformly dominant: at 3B IID, triples reach
the highest EM (0.956 vs.\ 0.952 for \texttt{hyper}), and sentences and
triples tie \texttt{hyper} on IID fact F1 (0.980 vs.\ 0.984). The clearest
advantage appears at 1.5B on OOD data, where \texttt{hyper} reaches EM
0.914 and fact F1 0.939---7.6 and 6.6 points above triples and well above
raw text (EM 0.603). Under distribution shift, degradation also differs:
at 1.5B, \texttt{raw} falls from 0.715 to 0.603 EM (IID$\to$OOD), whereas
\texttt{hyper} falls only from 0.936 to 0.914; \texttt{sentences} actually
exceeds \texttt{triples} on OOD EM at this scale (0.850 vs.\ 0.838), but
both remain below \texttt{hyper}.

A complementary pattern emerges for \emph{feasibility} detection
(Table~\ref{tab:feas}). Raw observations achieve the highest feasibility
accuracy (0.965 IID / 0.944 OOD at 1.5B) because environment text mentions
visible objects directly. Independent sentences and triples degrade
feasibility sharply at 0.5B ($\sim$0.72 accuracy) by scattering joint
preconditions across lines. \texttt{hyper} recovers much of this signal
(0.944 / 0.913 at 1.5B) while predicting effects more accurately than
fact-based flat or pairwise renderings in the same setting---a favorable
trade-off, though not the best on feasibility alone.

At 3B, all fact-based serializers improve substantially. \texttt{hyper}
achieves the best OOD fact F1 (0.975) and OOD EM (0.939), but IID EM favors
triples. This suggests that higher-capacity models partially compensate for
fragmented structure, narrowing---without eliminating---the benefit of
hyperedge grouping.

\begin{table}[t]
\centering
\small
\caption{World-model prediction on test-IID and test-OOD (effect exact match
/ fact F1). \textbf{Bold}: best value within each split and model scale.}
\label{tab:main}
\begin{tabular}{lcccccc}
\toprule
& \multicolumn{2}{c}{0.5B} & \multicolumn{2}{c}{1.5B} & \multicolumn{2}{c}{3B} \\
\cmidrule(lr){2-3}\cmidrule(lr){4-5}\cmidrule(lr){6-7}
Repr & EM & F1 & EM & F1 & EM & F1 \\
\midrule
\multicolumn{7}{l}{\textit{test-IID}} \\
Raw       & 0.710 & 0.853 & 0.715 & 0.861 & 0.726 & 0.871 \\
Sentences & 0.865 & 0.866 & 0.848 & 0.877 & 0.946 & 0.980 \\
Triples   & 0.889 & 0.892 & 0.856 & 0.878 & \textbf{0.956} & 0.980 \\
Hyper     & \textbf{0.943} & \textbf{0.963} & \textbf{0.936} & \textbf{0.956} & 0.952 & \textbf{0.984} \\
\midrule
\multicolumn{7}{l}{\textit{test-OOD}} \\
Raw       & 0.617 & 0.814 & 0.603 & 0.805 & 0.621 & 0.823 \\
Sentences & 0.784 & 0.811 & 0.850 & 0.873 & 0.913 & 0.952 \\
Triples   & 0.790 & 0.818 & 0.838 & 0.873 & 0.927 & 0.961 \\
Hyper     & \textbf{0.909} & \textbf{0.943} & \textbf{0.914} & \textbf{0.939} & \textbf{0.939} & \textbf{0.975} \\
\bottomrule
\end{tabular}
\end{table}

\begin{table}[t]
\centering
\small
\caption{Feasibility detection accuracy (1.5B). Raw is best; \texttt{hyper}
is second and substantially above sentences/triples.}
\label{tab:feas}
\begin{tabular}{lcc}
\toprule
Repr & test-IID & test-OOD \\
\midrule
Raw       & 0.965 & 0.944 \\
Sentences & 0.923 & 0.874 \\
Triples   & 0.922 & 0.887 \\
\textbf{Hyper} & 0.944 & 0.913 \\
\bottomrule
\end{tabular}
\end{table}

Three random seeds at 1.5B confirm stability: OOD EM for \texttt{hyper}
is $0.914 \pm 0.004$ (mean$\pm$std), versus $0.826 \pm 0.073$ for
sentences and $0.805 \pm 0.038$ for triples.

\subsection{Sample Efficiency}
\label{sec:results-sample}

\begin{figure}[t]
\centering
\includegraphics[width=0.75\linewidth]{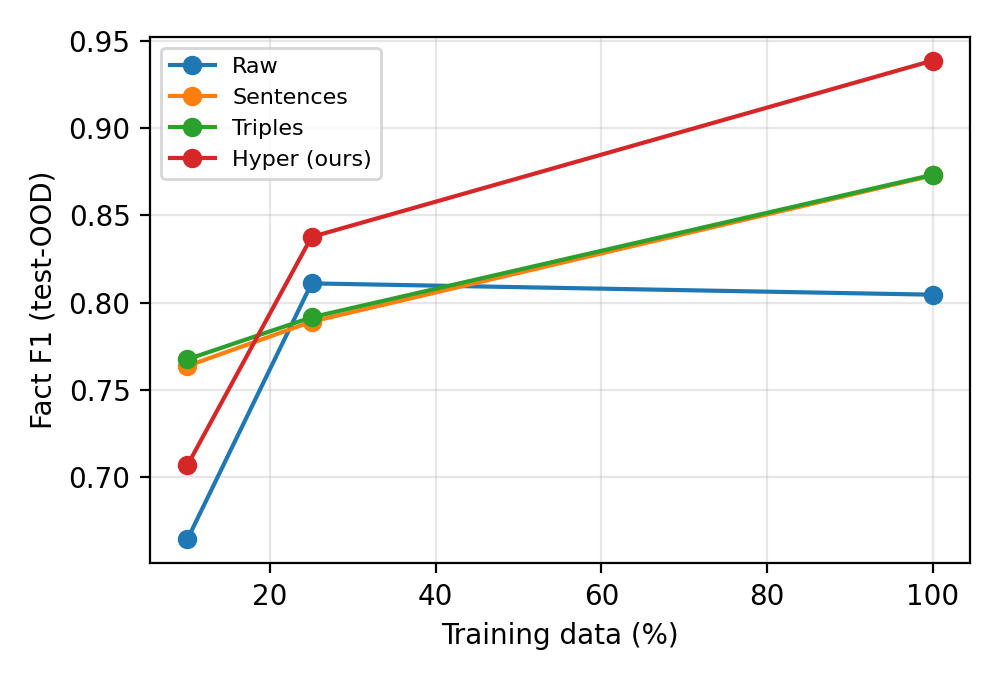}
\caption{OOD fact F1 vs.\ training data fraction (1.5B). \texttt{hyper}
leads on OOD from 25\% data upward; at 10\% all methods are close.}
\label{fig:sample}
\end{figure}

Figure~\ref{fig:sample} varies the training fraction at 1.5B. With only
10\% of training data, all representations perform similarly on OOD EM
($\sim$0.63--0.64), so structure alone is insufficient in the extreme
low-data regime. With 25\% data, \texttt{hyper} reaches OOD EM 0.824,
clearly above triples (0.703) and sentences (0.709); on IID EM at the same
fraction, however, sentences (0.885) still exceed \texttt{hyper} (0.836).
Hyperedge grouping therefore helps most on harder OOD generalization once a
modest amount of training data is available, but does not dominate every
metric at every budget.

\subsection{Multi-Step Rollouts}
\label{sec:results-rollout}

\begin{figure}[t]
\centering
\includegraphics[width=\linewidth]{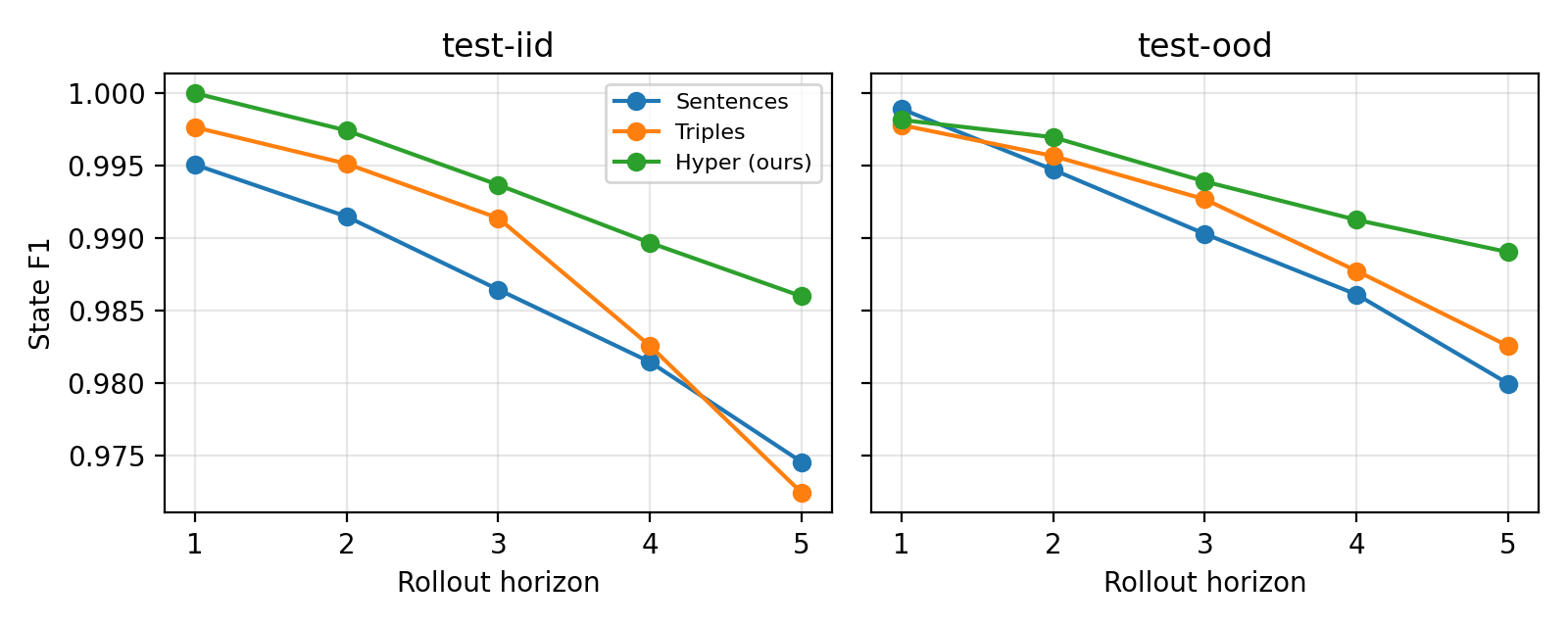}
\caption{Multi-step rollout state F1 vs.\ horizon (1.5B). Left: test-IID;
right: test-OOD.}
\label{fig:rollout}
\end{figure}

We apply predicted deltas iteratively along recorded trajectories and measure
state-set F1 at horizons 1--5. Figure~\ref{fig:rollout} shows modest but consistent rollout gains for
\texttt{hyper} at 1.5B: at horizon 5 on test-OOD, state F1 is 0.989 for
\texttt{hyper} versus 0.980 for sentences and 0.983 for triples. Better
single-step prediction therefore carries over, albeit with small margins, to
multi-step imagination.

\subsection{World-Model-Guided Planning}
\label{sec:results-planning}

Table~\ref{tab:plan} evaluates the learned world models inside a greedy
planner on 30 held-out test-IID games. With the default scoring rule,
\texttt{hyper} attains 76.7\% success, clearly above random (23.3\%),
sentences (53.3\%), and triples (56.7\%), and uses fewer steps on average
(14.3 vs.\ 22.9--34.3). Better learned dynamics therefore transfer to
downstream task completion under matched search.

Removing the confidence term $w_c\,c(a)$ in Eq.~\eqref{eq:score} raises
\texttt{hyper} success further to 93.3\% (9.9 steps), indicating that the
default confidence weight is miscalibrated rather than helpful. We report
both settings for transparency and leave calibrated confidence--planning
integration to future work.

\begin{table}[t]
\centering
\small
\caption{World-model-guided planning on 30 test-IID games (1.5B world
models).}
\label{tab:plan}
\begin{tabular}{lccc}
\toprule
Planner & WM repr & Success & Avg.\ steps \\
\midrule
Random            & ---        & 23.3\% & 34.3 \\
WM-guided         & sentences  & 53.3\% & 22.9 \\
WM-guided         & triples    & 56.7\% & 20.5 \\
WM-guided         & \textbf{hyper} & \textbf{76.7\%} & \textbf{14.3} \\
WM-guided (no conf.) & hyper   & 93.3\% &  9.9 \\
\bottomrule
\end{tabular}
\end{table}

\subsection{Case Study}
\label{sec:case}

Consider a test game requiring: \emph{take rusty key}, \emph{unlock chest
with rusty key}, \emph{take gold coin}. In the initial state the planner
must choose among admissible commands including \texttt{open chest} (infeasible
while locked) and \texttt{go east} (irrelevant).

The \texttt{raw} world model often predicts feasible effects for
\texttt{open chest} because the observation mentions the chest, yielding
misleading positive imaginations. The \texttt{triples} model represents
\texttt{locked(chest)} and \texttt{match(rusty key, chest)} on separate
lines; it may predict partial unlock effects without jointly satisfying all
preconditions. The \texttt{hyper} model renders
\texttt{[chest | state: closed, locked | key: rusty key]} as one unit and
correctly predicts \texttt{INFEASIBLE} for \texttt{open chest} and
\texttt{ADD: none | REMOVE: locked(chest)} for \texttt{unlock chest with
rusty key}, guiding the planner along the shortest successful path in 10
steps rather than exhausting the 40-step budget.

\section{Conclusion}
\label{sec:conclusion}

We presented \method, a controlled study of how state-serialization
structure affects learned textual world models. Hyperedge grouping is not
uniformly optimal---at 3B IID, triples match or exceed it on EM---but it
offers the strongest overall profile at 0.5B--1.5B, the best OOD
generalization in most settings, near-raw feasibility at 1.5B, and the
highest planning success among matched world models (76.7\% vs.\ 53--57\%
for flat/pairwise alternatives). On TextWorld, this yields up to 31
percentage-point OOD EM gains over raw text at 1.5B. More broadly, the
result connects state-serialization design with structured prediction in
non-stationary dynamical systems, where recent spatio-temporal graph
attention models use graph structure to capture evolving dependencies
among interacting variables \citep{wei2025fstgat}. It also suggests a
possible bridge to long-horizon vision--language--action manipulation, in
which symmetry-aware decision-making and structured state abstraction are
important for planning over extended action sequences \citep{jian2026pi}.
Future work includes learned hyperedge grouping, calibrated confidence for
planning, and extension to richer environments such as ALFWorld,
ScienceWorld, and embodied VLA manipulation tasks.

\bibliographystyle{unsrtnat}
\bibliography{references}

@article{ha2018world,
  title={World models},
  author={Ha, David and Schmidhuber, J{\"u}rgen},
  journal={arXiv preprint arXiv:1803.10122},
  year={2018}
}

@article{hafner2023dreamerv3,
  title={Mastering diverse domains through world models},
  author={Hafner, Danijar and Pasukonis, Jurgis and Ba, Jimmy and Lillicrap, Timothy},
  journal={arXiv preprint arXiv:2301.04104},
  year={2023}
}

@misc{lecun2022path,
  title={A path towards autonomous machine intelligence},
  author={LeCun, Yann},
  howpublished={OpenReview},
  year={2022},
  url={https://openreview.net/forum?id=BZ5a1r-kVsf}
}

@inproceedings{bruce2024genie,
  title={Genie: Generative interactive environments},
  author={Bruce, Jake and Dennis, Michael and Edwards, Ashley and others},
  booktitle={International Conference on Machine Learning},
  year={2024}
}

@article{agarwal2025cosmos,
  title={Cosmos world foundation model platform for physical {AI}},
  author={Agarwal, Niket and Ali, Arslan and Bala, Maciej and others},
  journal={arXiv preprint arXiv:2501.03575},
  year={2025}
}

@inproceedings{ammanabrolu2021worldformer,
  title={Learning knowledge graph-based world models of textual environments},
  author={Ammanabrolu, Prithviraj and Riedl, Mark},
  booktitle={Advances in Neural Information Processing Systems},
  volume={34},
  year={2021}
}

@article{anokhin2024arigraph,
  title={{AriGraph}: Learning knowledge graph world models with episodic memory for {LLM} agents},
  author={Anokhin, Petr and Semenov, Nikita and Sorokin, Artyom and Evseev, Dmitry and Kravchenko, Andrey and Burtsev, Mikhail and Burnaev, Evgeny},
  journal={arXiv preprint arXiv:2407.04363},
  year={2024},
  note={Published at IJCAI 2025}
}

@article{xu2025gwm,
  title={Graph world model},
  author={Feng, Tao and Wu, Yexin and Lin, Guanyu and You, Jiaxuan},
  journal={arXiv preprint arXiv:2507.10539},
  year={2025}
}

@article{statefactory2026,
  title={Reward prediction with factorized world states},
  author={Shen, Yijun and Chen, Delong and Hu, Xianming and Mi, Jiaming and Zhao, Hongbo and Zhang, Kai and Fung, Pascale},
  journal={arXiv preprint arXiv:2603.09400},
  year={2026}
}

@inproceedings{chae2025web,
  title={Web agents with world models: Learning and leveraging environment dynamics in web navigation},
  author={Chae, Hyungjoo and Kim, Namyoung and Ong, Kai Tzu-iunn and Gwak, Minju and Song, Gwanwoo and Kim, Jihoon and Kim, Sunghwan and Lee, Dongha and Yeo, Jinyoung},
  booktitle={International Conference on Learning Representations},
  year={2025}
}

@article{gu2024proposes,
  title={Is your {LLM} secretly a world model of the internet? Model-based planning for web agents},
  author={Gu, Yu and Zheng, Boyuan and Gou, Boyu and Zhang, Kai and Chang, Cheng and Srivastava, Sanjari and Xie, Yanan and Qi, Peng and Sun, Huan and Su, Yu},
  journal={arXiv preprint arXiv:2411.06559},
  year={2024}
}

@article{liu2026itp,
  title={Imagine-then-plan: Agent learning from adaptive lookahead with world models},
  author={Liu, Youwei and Wang, Jian and Wang, Hanlin and Guo, Beichen and Li, Wenjie},
  journal={arXiv preprint arXiv:2601.08955},
  year={2026}
}

@inproceedings{luo2025hypergraphrag,
  title={{HyperGraphRAG}: Retrieval-augmented generation via hypergraph-structured knowledge representation},
  author={Luo, Haoran and E, Haihong and Chen, Guanting and Zheng, Yandan and Wu, Xiaobao and Guo, Yikai and Lin, Qika and Feng, Yu and Kuang, Zemin and Song, Meina and Zhu, Yifan and Luu, Anh Tuan},
  booktitle={Advances in Neural Information Processing Systems},
  volume={39},
  year={2025}
}

@inproceedings{cote2018textworld,
  title={{TextWorld}: A learning environment for text-based games},
  author={C{\^o}t{\'e}, Marc-Alexandre and K{\'a}d{\'a}r, {\'A}kos and Yuan, Xingdi and others},
  booktitle={Workshop on Computer Games},
  pages={41--75},
  year={2018},
  organization={Springer}
}

@article{qwen2.5,
  title={Qwen2.5 technical report},
  author={Yang, An and Yang, Baosong and Zhang, Beichen and others},
  journal={arXiv preprint arXiv:2412.15115},
  year={2024}
}

@inproceedings{hu2022lora,
  title={{LoRA}: Low-rank adaptation of large language models},
  author={Hu, Edward J and Shen, Yelong and Wallis, Phillip and others},
  booktitle={International Conference on Learning Representations},
  year={2022}
}

@article{assran2025vjepa2,
  title={{V-JEPA 2}: Self-supervised video models enable understanding, prediction and planning},
  author={Assran, Mahmoud and Bardes, Adrien and Fan, David and Garrido, Quentin and Howes, Russell and Komeili, Mojtaba and Muckley, Matthew and Rizvi, Ammar and Roberts, Claire and Sinha, Koustuv and Zholus, Artem and others},
  journal={arXiv preprint arXiv:2506.09985},
  year={2025}
}

@inproceedings{zhao2017algorithm,
  title={An algorithm for the orientation of complete bipartite graphs},
  author={Zhao, Lingqi and Wang, Mujiangshan and Zhang, Xuefei and Lin, Yuqing and Wang, Shiying},
  booktitle={2017 International Conference on Applied Mathematics, Modelling and Statistics Application (AMMSA 2017)},
  pages={361--364},
  year={2017},
  organization={Atlantis Press}
}

@article{wang2016diagnosability,
  title={Diagnosability of Cayley graph networks generated by transposition trees under the comparison diagnosis model},
  author={Wang, Mujiangshan and Wang, Shiying},
  journal={Annals of Applied Mathematics},
  volume={32},
  number={2},
  pages={166--173},
  year={2016}
}

@article{wang2018edge,
  title={The Edge Connectivity of Expanded k-Ary n-Cubes},
  author={Wang, Shiying and Wang, Mujiangshan},
  journal={Discrete Dynamics in Nature and Society},
  volume={2018},
  number={1},
  pages={7867342},
  year={2018},
  publisher={Wiley Online Library}
}

@article{mu2010ordered,
  title={Ordered and Hamilton Digraphs},
  author={Mu-Jiang-shan, WANG and Jun, YUAN and Shang-wei, LIN and others},
  journal={Chinese Quarterly Journal of Mathematics},
  volume={25},
  number={3},
  pages={317--326},
  year={2010}
}

@article{wei2025fstgat,
  title={FSTGAT: Financial Spatio-Temporal Graph Attention Network for Non-Stationary Financial Systems and Its Application in Stock Price Prediction},
  author={Wei, Ze-Lin and An, Hong-Yu and Yao, Yao and Su, Wei-Cong and Li, Guo and Saifullah and Sun, Bi-Feng and Wang, Mu-Jiang-Shan},
  journal={Symmetry},
  volume={17},
  number={8},
  pages={1344},
  year={2025},
  publisher={MDPI}
}

@inproceedings{wang2011embedding,
  title={Embedding paths into the 4-ary n-cube with faulty nodes},
  author={Wang, Shiying and Wangmu, Jiangshan and Qi, Zhifang and Ren, Yunxia},
  booktitle={2011 International Conference on Consumer Electronics, Communications and Networks (CECNet)},
  pages={4949--4951},
  year={2011},
  organization={IEEE}
}

@article{jian2026pi,
  title={PI-VLA: Adaptive Symmetry-Aware Decision-Making for Long-Horizon Vision--Language--Action Manipulation},
  author={Jian, Yina and Tian, Di and Chen, Xuan-Jing and Wei, Zhen-Yuan and Liang, Chen-Wei and Wang, Mu-Jiang-Shan},
  journal={Symmetry},
  volume={18},
  number={3},
  pages={394},
  year={2026},
  publisher={MDPI}
}

\end{document}